\documentclass[10pt,twocolumn,lethttps://www.overleaf.com/7547131597gqfwfzhdkdqbterpaper]{article}

\usepackage{cvpr}
\usepackage{times}
\usepackage{epsfig}
\usepackage{graphicx}
\usepackage{amsmath}
\usepackage{amssymb}
\usepackage{wrapfig}
\usepackage{subfigure}
\usepackage{booktabs,caption}
\usepackage[flushleft]{threeparttable}
\usepackage{gensymb}
\usepackage{bm}
\usepackage{textpos}

\usepackage[pagebackref=true,breaklinks=true,letterpaper=true,colorlinks,bookmarks=false]{hyperref}

\newcommand{\msplit}{\mathcal{M}}

\cvprfinalcopy 

\def\cvprPaperID{0027} 

\begin{document}
	
\begin{textblock*}{\textwidth}(0cm,0cm)
	\large\noindent{\copyright~2019 IEEE.  Personal use of this material is permitted.  Permission from IEEE must be obtained for all other uses, in any current or future media, including reprinting/republishing this material for advertising or promotional purposes, creating new collective works, for resale or redistribution to servers or lists, or reuse of any copyrighted component of this work in other works.}
\end{textblock*}
\thispagestyle{empty}
	
\title{Bean Split Ratio for Dry Bean Canning Quality and Variety Analysis}

\author{Yunfei Long$^1$, Amber Bassett$^2$, Karen Cichy$^{2,3}$, Addie Thompson$^2$, and Daniel Morris$^1$\\\\		
$^1$Department of Electrical and Computer Engineering, Michigan State University\\
$^2$Department of Plant, Soil and Microbial Sciences, Michigan State University\\
$^3$Sugarbeet and Bean Research Unit, USDA-ARS\\
{\tt\small \{longyunf, basset31, thom1718, dmorris\}@msu.edu,~karen.cichy@ars.usda.gov}
}

\maketitle
\pagenumbering{arabic}

\begin{abstract}
Splits on canned beans appear in the process of preparation and canning. Researchers are studying how they are influenced by cooking environment and genotype. However, there is no existing method to automatically quantify or to characterize the severity of splits. To solve this, we propose two measures: the Bean Split Ratio (BSR) that quantifies the overall severity of splits, and the Bean Split Histogram (BSH) that characterizes the size distribution of splits.  We create a pixel-wise segmentation method to automatically estimate these measures from images.  We also present a bean dataset of recombinant inbred lines of two genotypes, use the BSR and BSH to assess canning quality, and explore heritability of these properties.  
\end{abstract}

\section{Introduction}

Beans are a staple food, having low cost and high nutrition (including protein, vitamins, minerals, and dietary fiber)~\cite{kutovs2003dietary}, and additional health benefits~\cite{zanovec2011comparison}. For quickness of use, beans are often pre-cooked and canned.  However, depending on genotype, bean structural integrity can be harmed by cooking and canning, so there is an effort to improve these traits through breeding ~\cite{Wassimi1990,lu1996correlations,Cichy2014}.  To facilitate this, automated quality measures are needed, and that is the focus of this paper. 

\begin{figure}[t]
	\centering
	\includegraphics[width=0.95\linewidth]{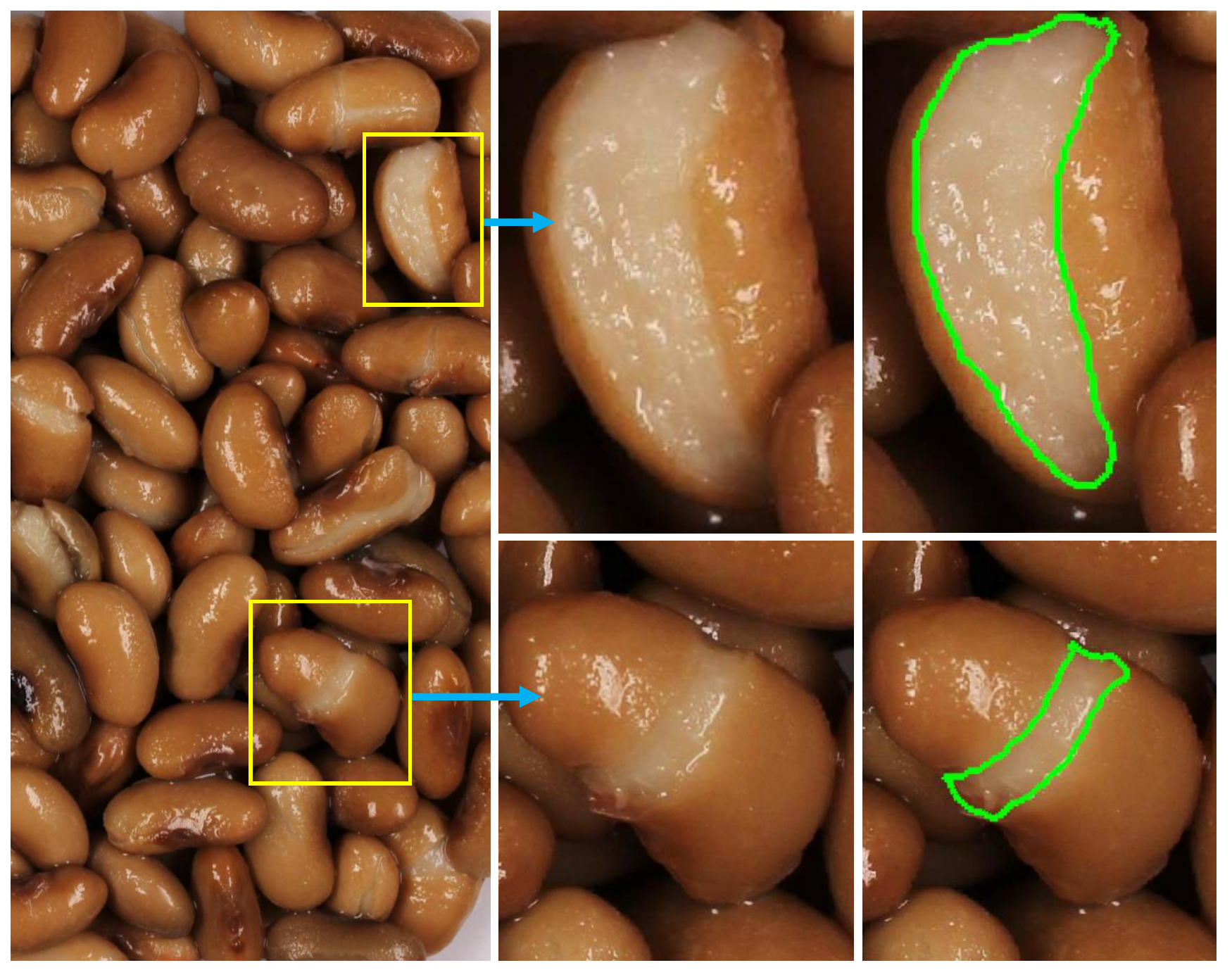}
	\caption{Examples of splits on canned yellow dry beans, which we propose to characterize with BSR and BSH. Detected splits are shown as green boundaries.}
	\label{Figure:split}
\end{figure}

Splits in the seed coats of canned dry beans, as shown in Fig.~\ref{Figure:split}, are an important undesirable phenotype exhibited following preparation and retort processing ~\cite{mendoza2017automated,Wassimi1990,vanburen1986}. Preparation for canning involves soaking, which may reveal splits that are already present in the dry seeds or induce splitting in genotypes with poor seed coat integrity ~\cite{SCHOENINGER2017,vanburen1986,Zamindar2013}. Following preparation, beans are canned in a retort under high temperature, high pressure conditions that can further induce splitting ~\cite{SCHOENINGER2017,vanburen1986}. Splits degrade the appearance and internal quality of beans~\cite{mendoza2017automated,SCHOENINGER2017,vanburen1986}, and genotypes that exhibit extensive splitting following canning are considered to have poor canning quality. The canning industry demands beans that can withstand the canning process to produce a high quality product~\cite{SCHOENINGER2017,vanburen1986}, so it is important for dry bean breeders to evaluate canning quality of breeding lines prior to variety release. To improve the end-use quality of dry beans, researchers in plant and food science are interested in studying the association of splitting with genotype and processing conditions~\cite{Zamindar2013,Cichy2014} including cooking time and, in this case, retort processing time.

It is common practice to evaluate the appearance of canned beans by manual visual inspection prior to variety release~\cite{mendoza2017automated,DryBeanProduction}.  Trained reviewers rate the overall quality of canned products poured into trays, taking into account intactness, color, clumping, and brine characteristics including color, viscosity, and amount of extruded starch~\cite{hosfield1995eight}. This, however, is biased, imprecise, laborious, and time-consuming~\cite{mendoza2017automated}. To automate this process, Mendoza et. al.~\cite{mendoza2017automated} extracted hand-crafted features such as color and texture from bean images and used a support vector machine to classify canned bean samples into acceptable and unacceptable categories. Here our aim is to characterize and quantify only splits.  To the best of our knowledge, this is the first method that specializes in quantifying bean splits.

Characterizing splits involves identifying splits in images, but this is challenging, even for a human, due to variations in shape, appearance, and lighting.  Our approach leverages convolutional neural networks (ConvNet)~\cite{lecun1998gradient, lecun2015deep} which have been successful in image-based classification, detection, and segmentation, and play an increasingly important role  in agriculture and food science~\cite{kamilaris2018deep}. Object segmentation with ConvNets~\cite{long2015fully, guo2018review} achieve pixel-level classification for numerous applications from leaf segmentation~\cite{morris2018pyramid} to vehicle detection from Lidar points~\cite{li2016vehicle}.  Our bean split detector uses a ConvNet trained to discriminate splits from bean seed coats. 

In this work, we propose two new measures for quantifying and characterizing bean split severity.  We also present a tool to automatically estimate these measures from images and so characterize bean intactness. In addition, we establish a public bean-split dataset of canned bean images of different genotypes and retort processing times. Experimental results verify the feasibility of detecting and quantifying split severity and demonstrate the strategy's potential to be a powerful tool in revealing the association of splits with genotype and retort processing time.

\section{Related Work}

Various non-intrusive techniques such as soft X-ray imaging, hyperspectral imaging, and machine vision have been used to detect seed defects and evaluate seed quality~\cite{rahman2016assessment}, with machine vision having the advantage of being low-cost.  Efforts have been made to extract color features from images and use them to recognize damage types of soybeans~\cite{shatadal2003identifying} or to classify beans into a number of quality categories~\cite{mendoza2017automated, de2016computer}. However, the severity of the defects on beans has not been accurately represented and quantified.

Numerous techniques are used to reveal hidden defects in the process of imaging, making it possible to segment defect regions via simple image processing such as thresholding. For example, Sood~$\it{et~al.}$~\cite{sood2016internal} applied X-ray imaging to visualize internal cracks of kidney bean seeds. Momin~$\it{et~al.}$~\cite{momin2017machine} employed backlighting to enhance the splits of soybeans in captured images. In contrast, this work focuses on external splits of canned beans, and such splits are more complex and diverse in color and textures than that of uncooked soybean seeds.

Annotated image are the driving force behind the successful application of ConvNets to agriculture and food science. With large datasets, deep ConvNet achieves high performance in general tasks such as classification of plant species~\cite{dyrmann2016plant} and crop diseases~\cite{ferentinos2018deep, mohanty2016using} from images. In addition, researchers build smaller datasets for specific tasks, including quantifying plant phynotypes. For example, with a self-made dataset and applying ConvNet, Baweja~$\it{et~al.}$~\cite{baweja2018stalknet} measure stalk count and stalk width, and Pound~$\it{et~al.}$~\cite{pound2017deep} accurately localize wheat spikes and spikelets.  Here we create a small dataset for a specific bean phenotype task.

\section{Proposed Measures}

We seek to move beyond manual assessment of dry bean quality and build automated tools to quantify quality-related bean phenotypes.  For this goal, we propose two phenotypic measurements: Bean Split Ratio (BSR) and Bean Split Histogram (BSH), and present a tool for their automatic estimation.  This section describes these measurements.

\subsection{BSR: Bean Split Ratio}

We define the BSR as the ratio of the exposed split area of a bean to the total surface area of the bean and refer to it as $b$.  This will vary from bean to bean, so for a collection of beans the average BSR is:
\begin{eqnarray}
\label{eq:BSR}
b=\frac{A_S}{A_B},
\end{eqnarray}
where $A_S$ is the total split areas and  $A_B$ is the total bean surface areas. The BSR is also the probability that a bean surface point selected at random will be within a split. 

Now, given an image of beans in a tray, we can define an image-estimated BSR as the ratio of the total number of split pixels to the total number of bean pixels (including both bean seed coat and bean split pixels). If splits are uniformly distributed around beans, then each pixel has an independent probability of $b$ of being a split, and the expected value of the image-estimated BSR is $b$.   

The image-estimated BSR has a number of characteristics that make it useful and convenient.  It is simple and does not require segmenting or counting individual beans, which would involve additional labeling effort and classification.  It is robust to occlusions, which always exist as at most one half of a bean is observed in any single image, and occlusions become worse when beans overlap each other. Furthermore, the image-estimated BSR is invariant to scale and so does not require strict control of distance from camera or viewing angle.  Finally, BSR measures the lack of intactness of beans, a quantity used to assess bean quality, see Section~\ref{sect:dataset}.

\subsection{BSH: Bean Split Histogram}

In addition to their total area, splits can be characterized by their size distribution.  While one could enumerate the number of splits of different sizes, we instead propose a Bean Split Histogram (BSH) that sums the total area belonging to splits of different sizes, divided by the total bean area.  This gives more influence to larger splits and is a natural extension of the BSR.

\begin{figure*}[h]
	\centering
	\includegraphics[width=5in]{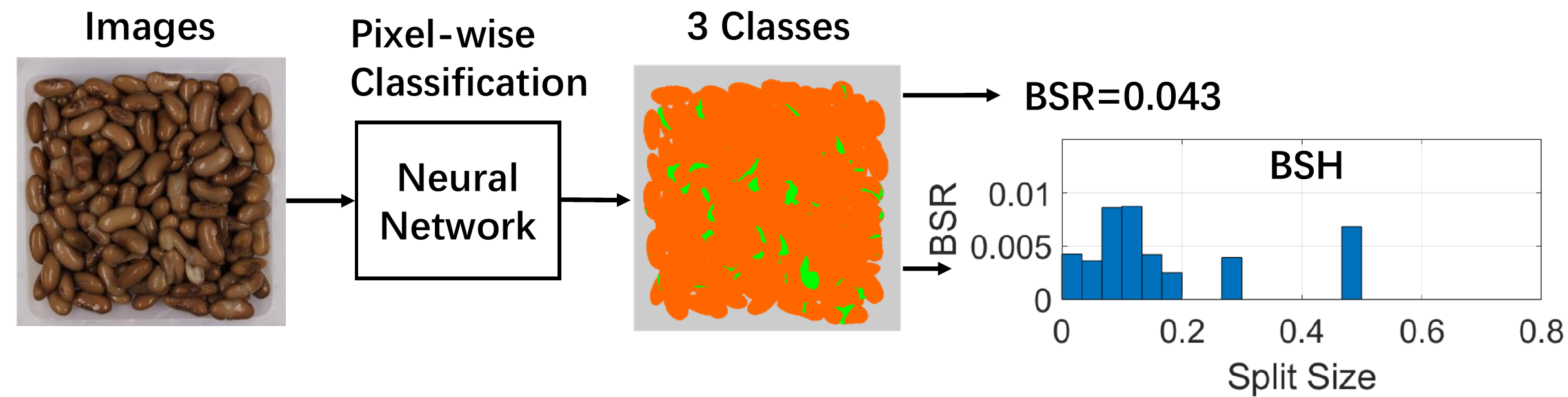}
	\caption{Overview of generating BSR and BSH from a bean image.}
	\label{Figure:diagram}
\end{figure*}

The BSH is a histogram of split area-ratios defined for a collection of beans.  Given $M$ splits, let $\{A_1,...,A_M\}$ refer to the areas of all the splits, and let $\msplit$ be the maximum split area (which could be equal to the maximum bean size).  Let the range $[0,1)$ be evenly partitioned into $N$ bins such that the $i$'th bin has range $R_i=[\frac{i-1}{N},\frac{i}{N})$.  Each bin of the BSH will collect the sum of BSR values from Eq.~\ref{eq:BSR} for splits of that size.  Thus the value, $b_i$, of the $i$'th bin of the BSH, is the sum of split ratios with area ratios in $R_i$, namely:
\begin{eqnarray}
b_i=\sum_{j=1}^M\delta^{(A_j)}_{ij}\frac{A_j}{A_B}, \hspace{0.5cm} \delta^{(A_j)}_{ij} = 
\begin{cases}
    1 & \mbox{if } \frac{A_j}{\msplit}\in R_i \\
    0 & \mbox{otherwise}.
\end{cases}
\label{eq:bsh}
\end{eqnarray}
An example BSH is shown in Fig.~\ref{Figure:diagram}.  A probabilistic interpretation of the BSH is that bin value $b_i$ is the probability of an point on the surface of the bean collection of being in a split whose area $A_j$ satisfies $\frac{A_j}{\msplit}\in R_i$.  We see that each split area contributes toward a single bin; therefore, the sum of bins is the BSR, namely $b=\sum_i^N b_i$.

In practice we use images to estimate a BSH.  For this the areas are replaced with image areas measured in pixels. We note that an image-BSH is not identical to the BSH in Eq.~ \ref{eq:bsh}.  The image-area of a split depends not only on its area but also on its orientation relative to the camera.  In addition, some splits may only be partially visible, and so be counted as a smaller split.  Also, if adjacent splits are merged during detection, they may be counted as a larger split.  However, given that most splits are small and dispersed, we expect that these adjustments are small and that the image-BSH will approximate the BSH.  

\section{Bean Dataset}
\label{sect:dataset}

A bean dataset\footnote{Dataset location: \url{https://www.egr.msu.edu/drybeans}} was collected and annotated for two purposes: (1) genotype and phenotype inheritance inference and (2) developing our split detection method.  The latter uses a smaller subset of the data, enabling us to generate BSR and BSH measures on the full dataset.  

\subsection{Dataset for genotype inferences}
\label{sec:genodataset}
The dry bean genotypes included in this study consist of two yellow dry bean parent genotypes and 18 of their recombinant inbred line progeny. The dataset includes a total of 20 genotypes that were processed in duplicate across five different retort processing times (10, 15, 20, 30 and 45 minutes) resulting in 200 canned samples. Human panelists performed quantitative visual ratings of canned samples to assess intactness. Further details of the bean genotypes, canning process, and quality evaluation are in the Appendix.

The beans were imaged as described in \cite{mendoza2017automated}. The system has a top-down view camera, fluorescent lights to provides stable lighting, and a cover to shield from ambient light. Each canned sample was rinsed to remove brine and then poured into a tray, positioned randomly and imaged.  A total of 199 images were collected from the 200 samples, as one sample was lost following visual evaluation. 

\subsection{Dataset for split detection}
\label{sec:splitdataset}
\begin{figure*}[h]
	\centering
	\subfigure[]
	{\label{Figure:lb1}
		\includegraphics[width=1.7in]{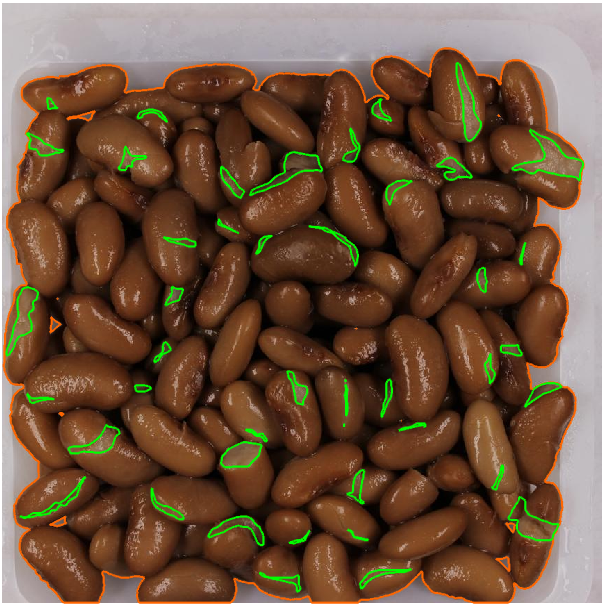}}
	\subfigure[]
	{\label{Figure:lb2}
		\includegraphics[width=1.7in]{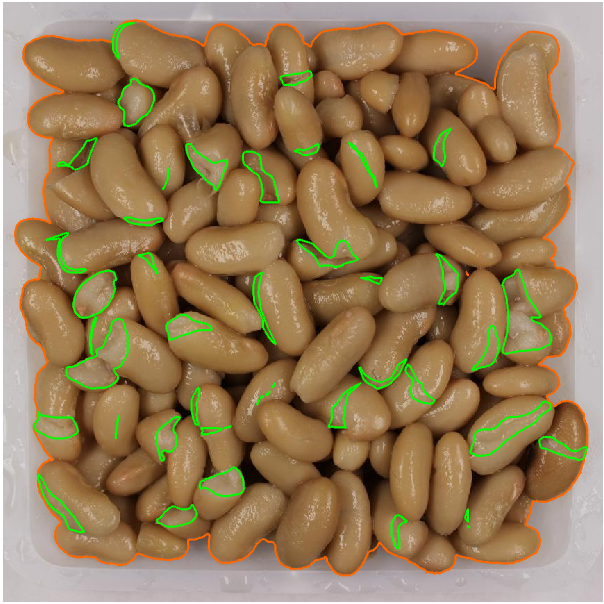}}
	\subfigure[]
	{\label{Figure:lb3}
		\includegraphics[width=1.7in]{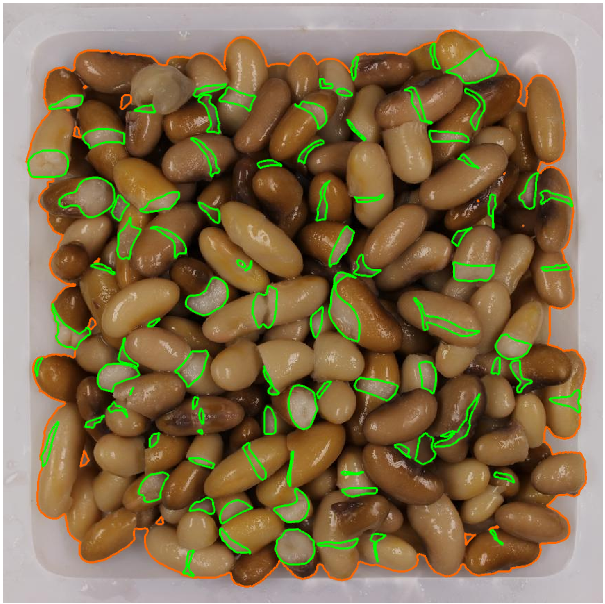}}
	\subfigure[]
	{\label{Figure:lb4}
		\includegraphics[width=1.7in]{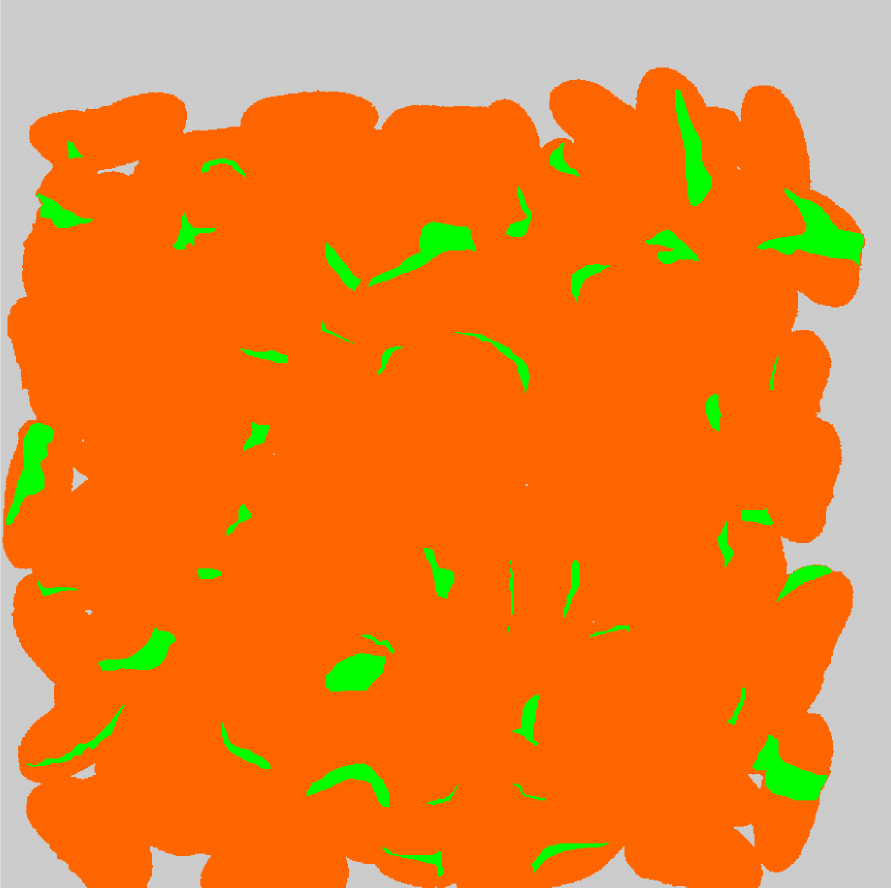}}
	\subfigure[]
	{\label{Figure:lb5}
		\includegraphics[width=1.7in]{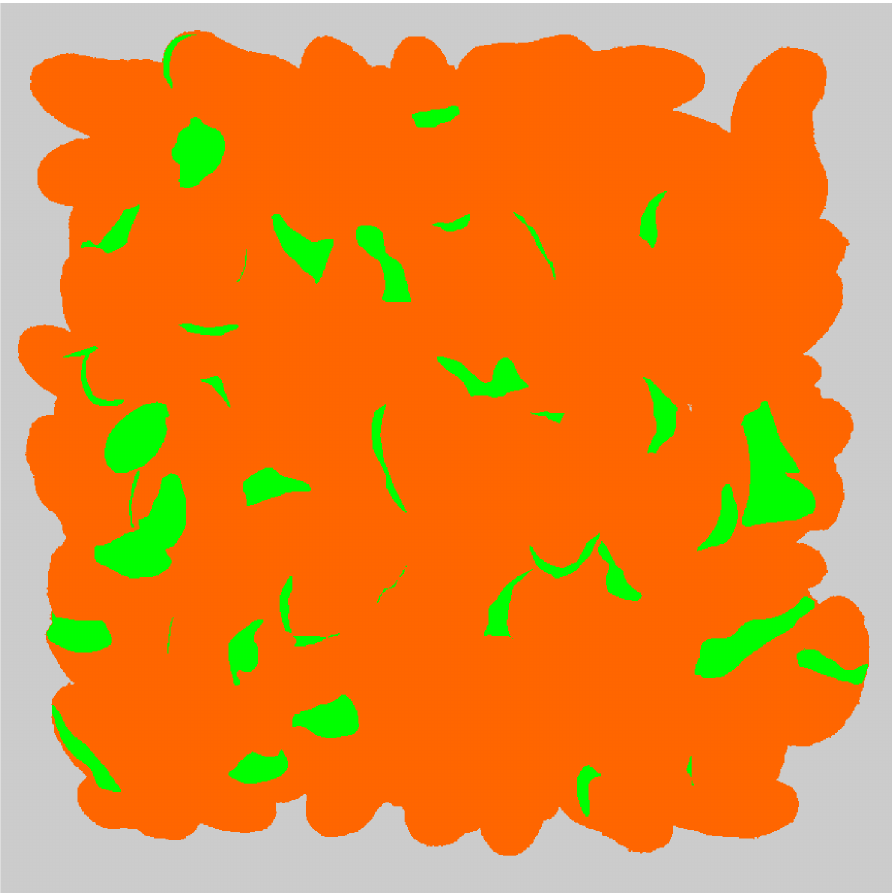}}
	\subfigure[]
	{\label{Figure:lb6}
		\includegraphics[width=1.7in]{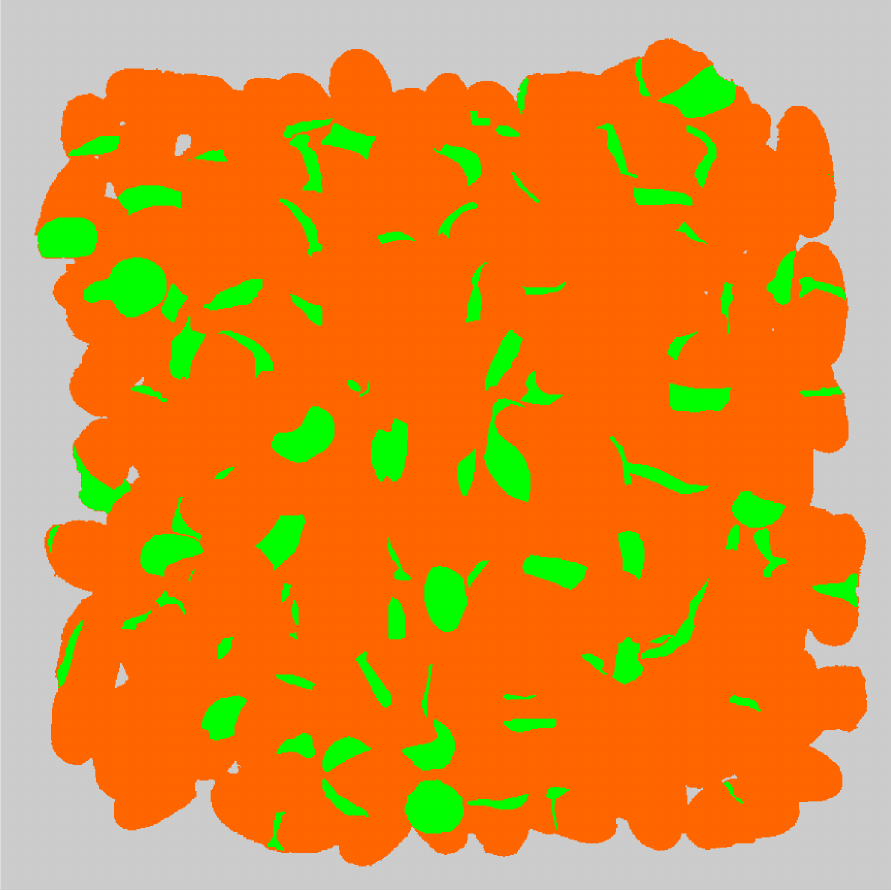}}
	\caption{Examples of pixel-wise annotations: (a) - (c) are representative images in our dataset with green lines between splits and seed coats, and orange between bean and tray pixels. (d) - (f) are visualizations of these same pixel-wise annotations with tray in gray, seed coat in orange, and splits in green.}
	\label{Figure:label}
\end{figure*}

A subset of 36 images from the dataset are pixel-wise labeled to enable development and validation of our split detection method, (29 for training and 7 for validation).  Each pixel is labeled as one of 3 classes: (1) tray, (2) seed coat, or (3) split.  Labeling was done using graph cuts~\cite{li2004lazy} to interactively separate bean pixels from tray pixels, and hand-drawn boundaries to segment each split.  Some examples are shown in Fig.~\ref{Figure:label}.

Although the number of labeled images is small, we believe it is feasible to train an accurate model from them for the following reasons. First, many genotypes have similar appearance that can be captured by a small number of images. Second, classification applies to a pixel, rather than image, and so each training image provides many labels. Third, images are augmented via rotations and flipping.

\section{Method}
We provide a summary of our method for estimating BSR and BSH from images as well as details of our segmentation algorithm.
\subsection{Algorithm}

\begin{enumerate}
	\item [A1:] Train two pixel-wise segmentation models using the labeled dataset described in Sec.~\ref{sec:splitdataset}.\vspace{-0.06in}
	
	\item [A2:] Apply trained models to generate pixel-wise, 3-class predictions on the full image dataset from Sec.~\ref{sec:genodataset}, namely: tray, seed coat, or split.\vspace{-0.06in}
	
	\item [A3:] Threshold and count the number of split pixels and seed coat pixels and compute the BSR using Eq.~\ref{eq:BSR}.\vspace{-0.06in}
	
	\item [A4:] Cluster split pixels into contiguous regions, measure their areas, and estimate the BSH using Eq.~\ref{eq:bsh}.\vspace{-0.06in}
\end{enumerate}

\subsection{Segmentation ConvNet}

Splits can be challenging to distinguish, even for humans, as surface textures and specular reflections often have similar appearance to splits, and some beans have seed coats with very similar color to the internal bean.  Discriminating splits from confusers involves using both fine texture cues and surrounding context.  This motivated our use of a pyramid convolutional network~\cite{morris2018pyramid} which integrates both fine detail and context to discriminate leaf boundaries.  

\subsubsection{Network architecture}
\label{sect:network}
\begin{figure*}[h]
	\centering
	\includegraphics[width= 6 in]{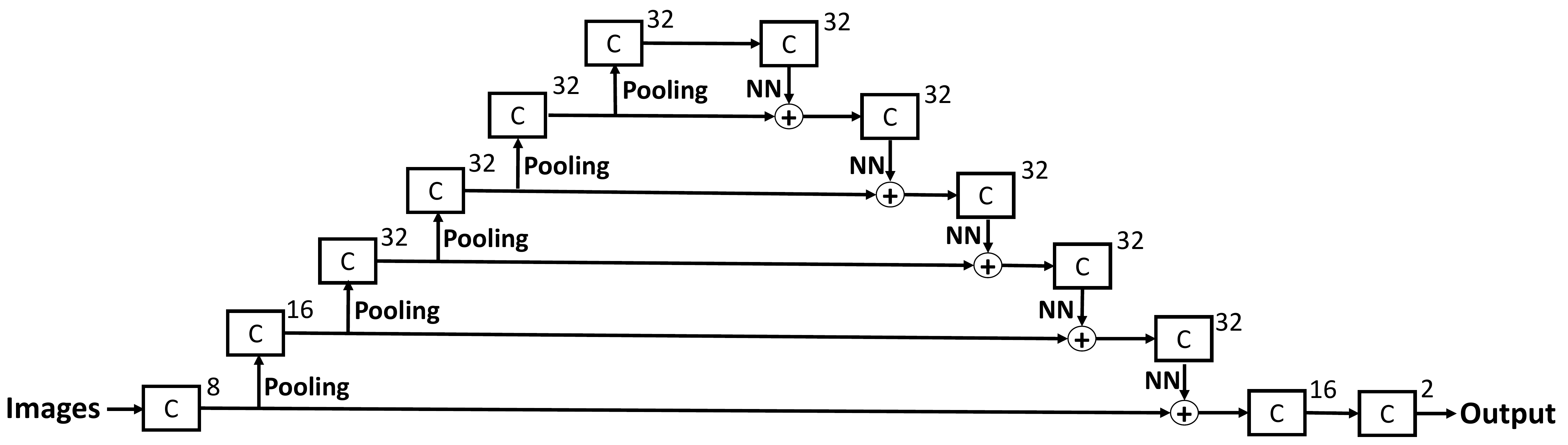}
	\caption{Network architecture: $C$ represents a convolutional layer using a $3\times3$ filter with $1\times1$ stride; the number of filter channels is shown after each box; NN denotes up-sampling via nearest neighbor interpolation.}
	\label{Figure:network}
\end{figure*}

The ConvNet architecture we use in this work is modified from~\cite{morris2018pyramid}, and its diagram is shown in Fig.~\ref{Figure:network}. Here symbol C represents sequential connection of convolutional layer and a ReLU nonlinearity, $\uparrow$ denotes down-sampling by max-pooling, and $\downarrow$ denotes up-sampling via nearest neighboring interpolation.  With six resolutions the network has a maximum receptive field size of $318\times318$, giving significant contextual cues for segmenting split pixels.

We train two networks independently: one performs bean/tray segmentation (orange + green vs. grey in Fig.~\ref{Figure:label}), and the other segments splits from seed coats (green vs. orange). The seed coat segmentation task is much more difficult than the former and requires more epochs for convergence, and so make sense to train separately.

\subsubsection{Thresholding scores}
\label{sect:threshold}
A threshold converts output scores into one of the 2 classes.  We compute the optimum threshold based on two criteria, (1) intersection over the union (IoU) between predicted split pixels with labeled pixels (a standard measure for detection accuracy) and (2) BSR error. Both give a similar best threshold level shown with the green lines.

\section{Experimental results}

Here we report on both the human panelists' rating and the automated BSR and BSH measures. We explore how these measures can be used to evaluate quality and investigate genetic factors influencing it.

\subsection{Canned bean intactness: human panelists}
Table~\ref{tab:variance} shows that the estimated intactness of canned bean samples was influenced by bean genotype, retort process time, and reviewer.  There was also an interaction between reviewer and genotype.  The effect of reviewer on the intactness score indicates the subjective nature of using human panelists for this evaluation, and the need for a more objective method, such as ours.   

\begin{table}
	\begin{threeparttable}
		\caption{Analysis of Variance of panel-estimated canned bean intactness.  This shows high variability of intactness due to genotype, retort time and reviewer.}
		\label{tab:variance}
		\begin{tabular}{llll}
			\toprule
			\textbf{Source$^1$} & \textbf{DF$^2$} & \textbf{F Value} & \textbf{PR$>$F} \\
			\midrule

			Genotype               & 19  & 30.51 & $<$.0001 \\
			Retort Time            & 4   & 29.92 & $<$.0001 \\
			Genotype by Retort Time   & 76  & 3.08  & $<$.0001 \\
			Replication (2 cans)   & 1   & 1.67  & $<$0.1968\\
			Reviewer               & 13  & 35.18 & $<$.0001 \\
			Genotype by Reviewer      & 247 & 1.63  & $<$.0001 \\
			\bottomrule
		\end{tabular}
		\begin{tablenotes}
			\small
			\item $^1$Type 3 tests in Proc Mixed with REML estimation method
			\item $^2$Numerator Degrees of Freedom for the F value calculation
		\end{tablenotes}
	\end{threeparttable}
\end{table}

\subsection{Image-based BSR prediction}
\begin{figure}[h]
	\centering
	\subfigure[]
	{\label{Figure:loss_bean}
		\includegraphics[width=2.8 in]{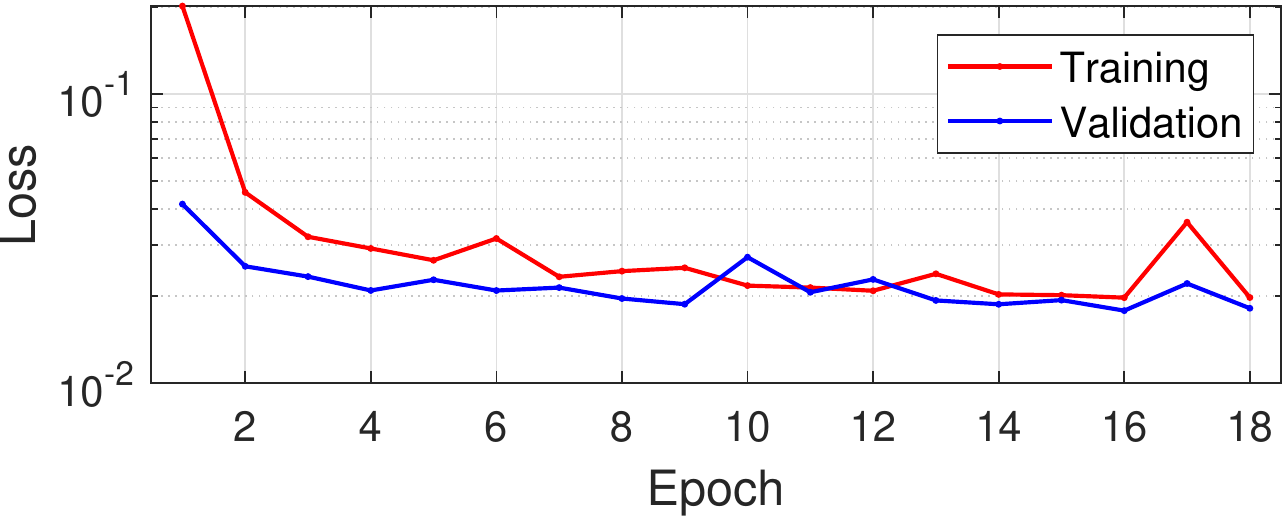}}
	\subfigure[]
	{\label{Figure:AP_bean}
		\includegraphics[width=2.8 in]{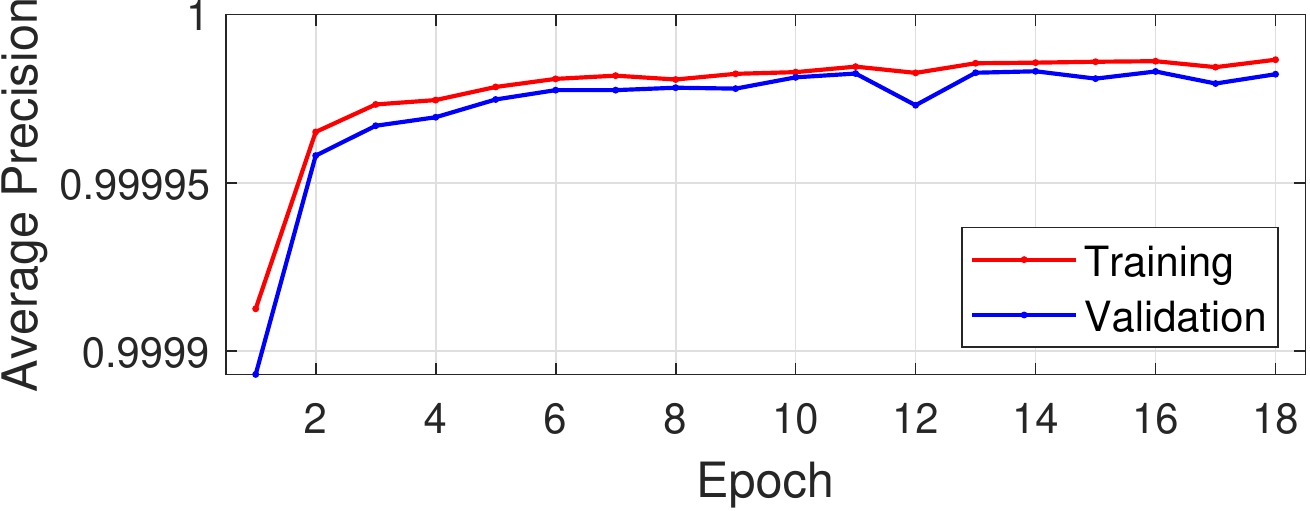}}
	\caption{Segmenting bean vs. tray is straightforward with our network and obtains high performance as shown by (a) Convergence of cross entropy loss, and (b) high average precision in pixel labels.}
	\label{Figure:bean_curve}
\end{figure}

We built the network shown in Fig.~\ref{Figure:network} based on Keras~\cite{chollet2015keras} with an input image resolution $864\times864$ and ran training programs with AdaDelta~\cite{zeiler2012adadelta} as an optimizer. As mentioned in Section~\ref{sect:network}, we trained two models. The first one distinguishes beans from the tray.  Loss and average precision (AP) over 18 epochs are shown in Figs.~\ref{Figure:loss_bean} and~\ref{Figure:AP_bean}, respectively. As it is an easy task, the loss function converges fast, and the model achieves high AP.

\begin{figure}[h]
	\centering
	\subfigure[]
	{\label{Figure:loss_split}
		\includegraphics[width=2.8 in]{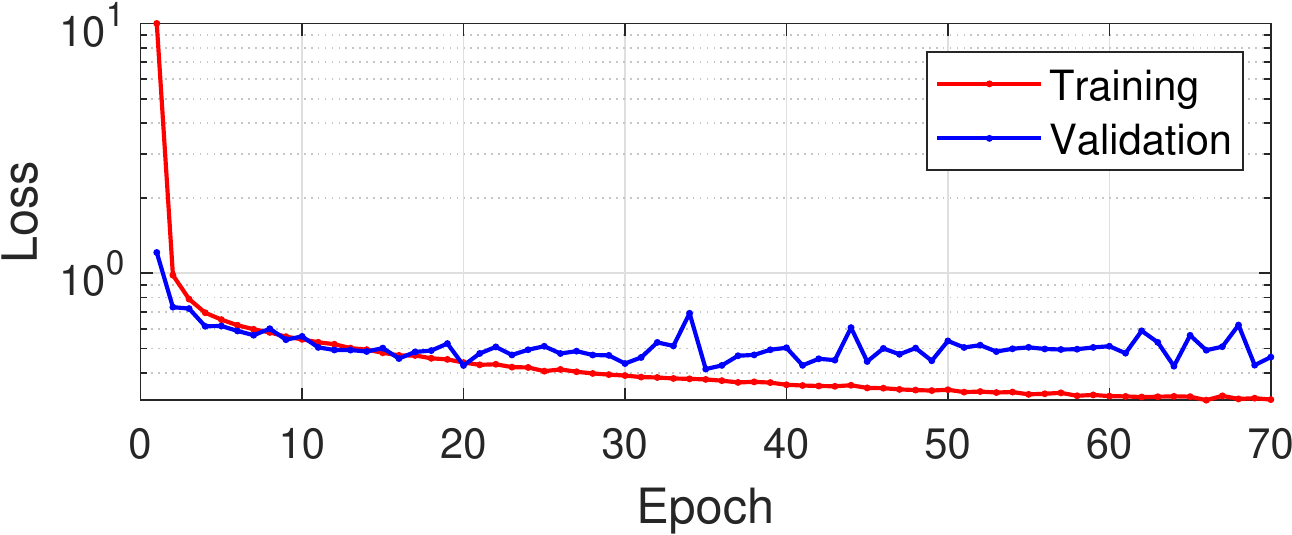}}
	\subfigure[]
	{\label{Figure:AP_split}
		\includegraphics[width=2.8 in]{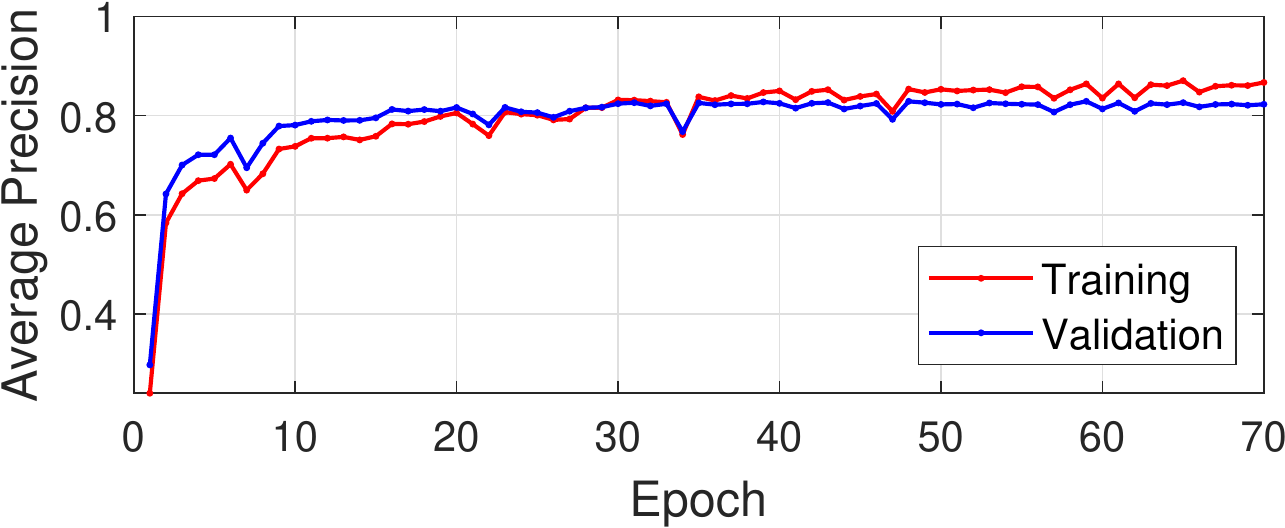}}
	\caption{Segmenting splits from seed coats is challenging.  (a) Loss and (b) average precision of pixel scores are shown.}
	\label{Figure:split_curve}
\end{figure}

The second model detects splits, by classifying each pixel into splits or seed coat. The tray pixels are ignored for computing loss and AP. The loss and AP curves over 70 epochs are shown in Figs.~\ref{Figure:loss_split} and~\ref{Figure:AP_split}, respectively. The model achieves AP over 0.8.

\begin{figure}[h]
	\centering
	\subfigure[]
	{\label{Figure:iou-T}
		\includegraphics[width=3 in]{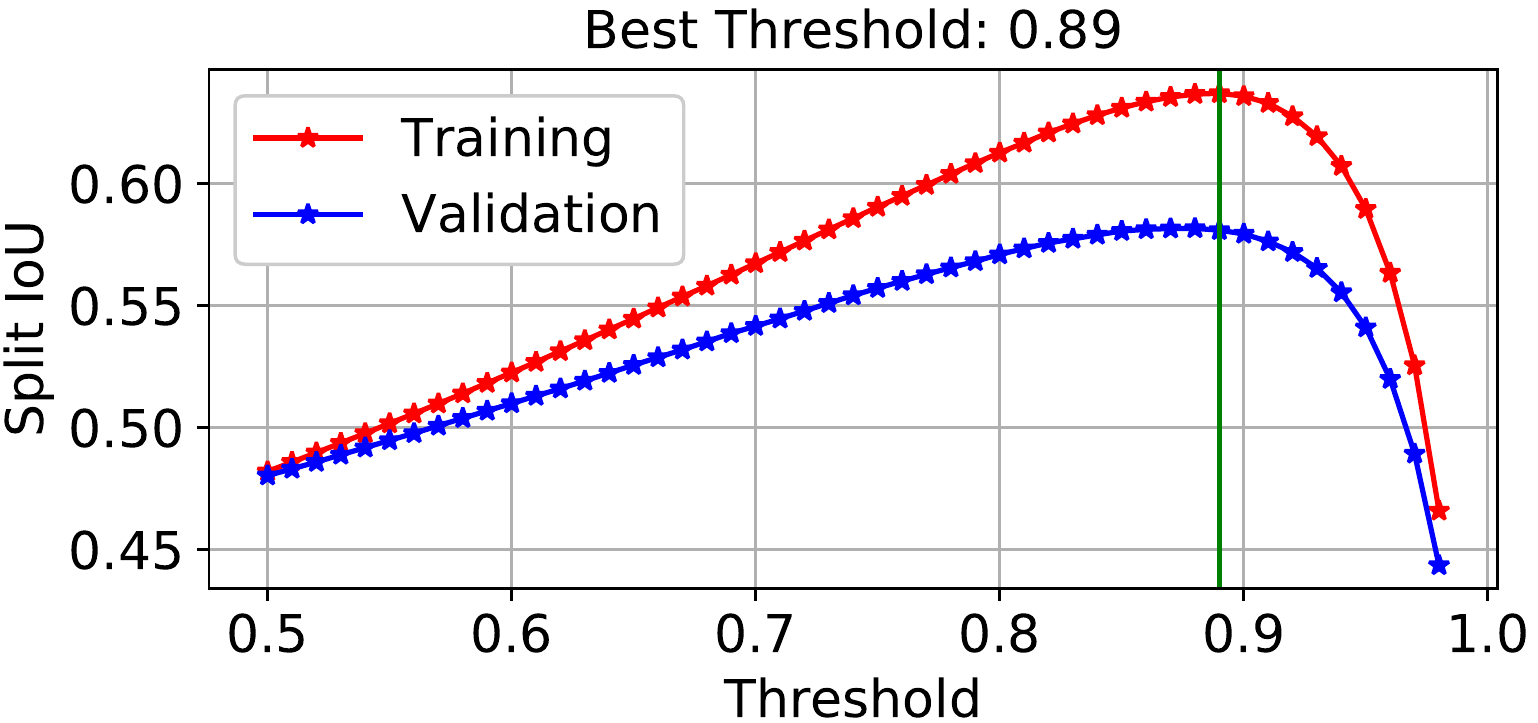}}
	\subfigure[]
	{\label{Figure:bsr-T}
		\includegraphics[width=3 in]{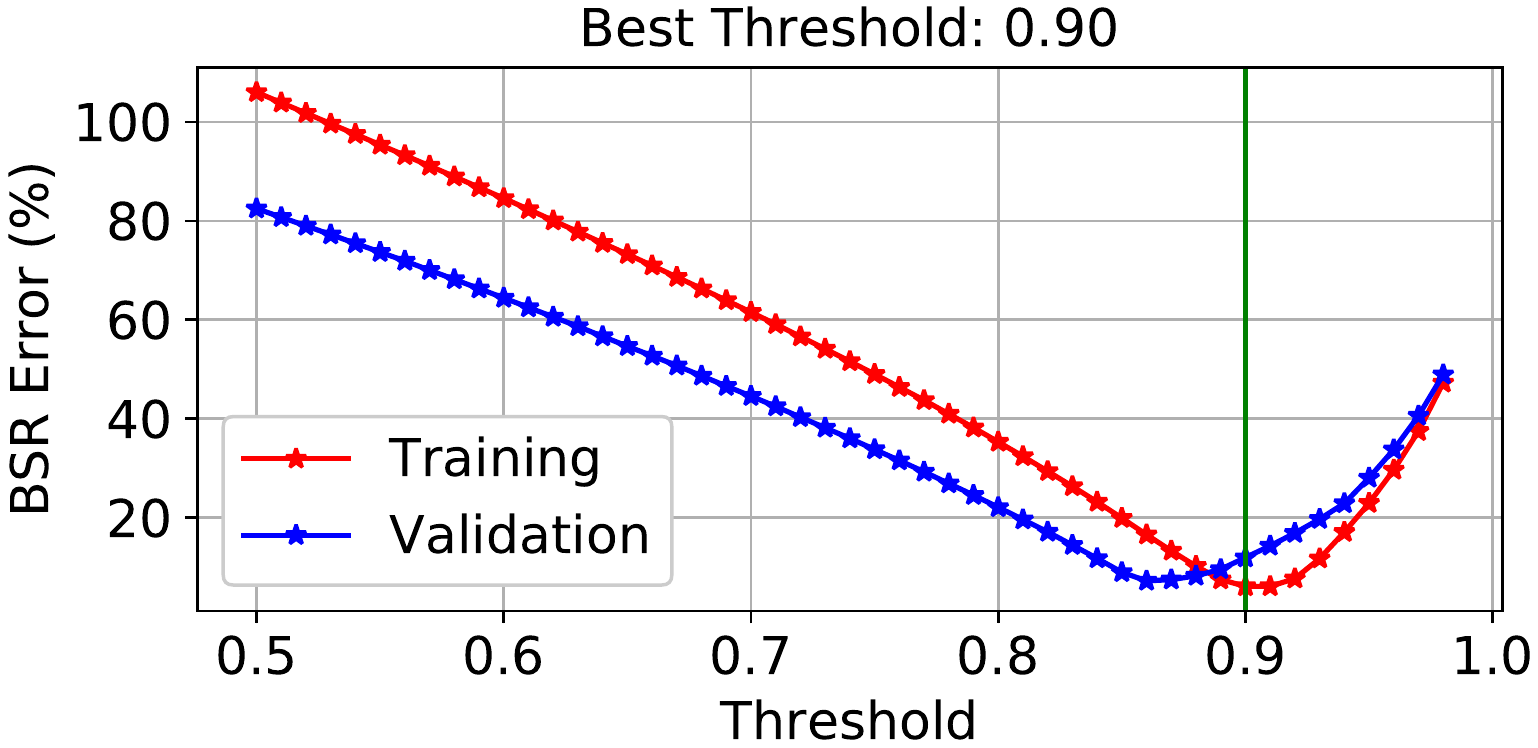}}
	\caption{We select a threshold (green line) that maximizes pixel segmentation performance measured as (a) the IoU of estimated split pixels, or (b) BSR percentage error.}
	\label{Figure:best_T}
\end{figure}

To use the predicted split pixels for further analysis, we need to apply a threshold on the network output.  It makes sense to select a threshold that maximizes performance, and Figs.~\ref{Figure:iou-T} and~\ref{Figure:bsr-T} show performance on two measures: Split IoU and BSR error. BSR error is the average percent error in estimating the BSR from an image.  It can be seen that the two criteria result in similar best threshold: $\it{i.e.}$ 0.89 (based on IoU) and 0.90 (based on BSR error).

\begin{figure}[ht]
    \begin{center}
    \includegraphics[width=2.8 in]{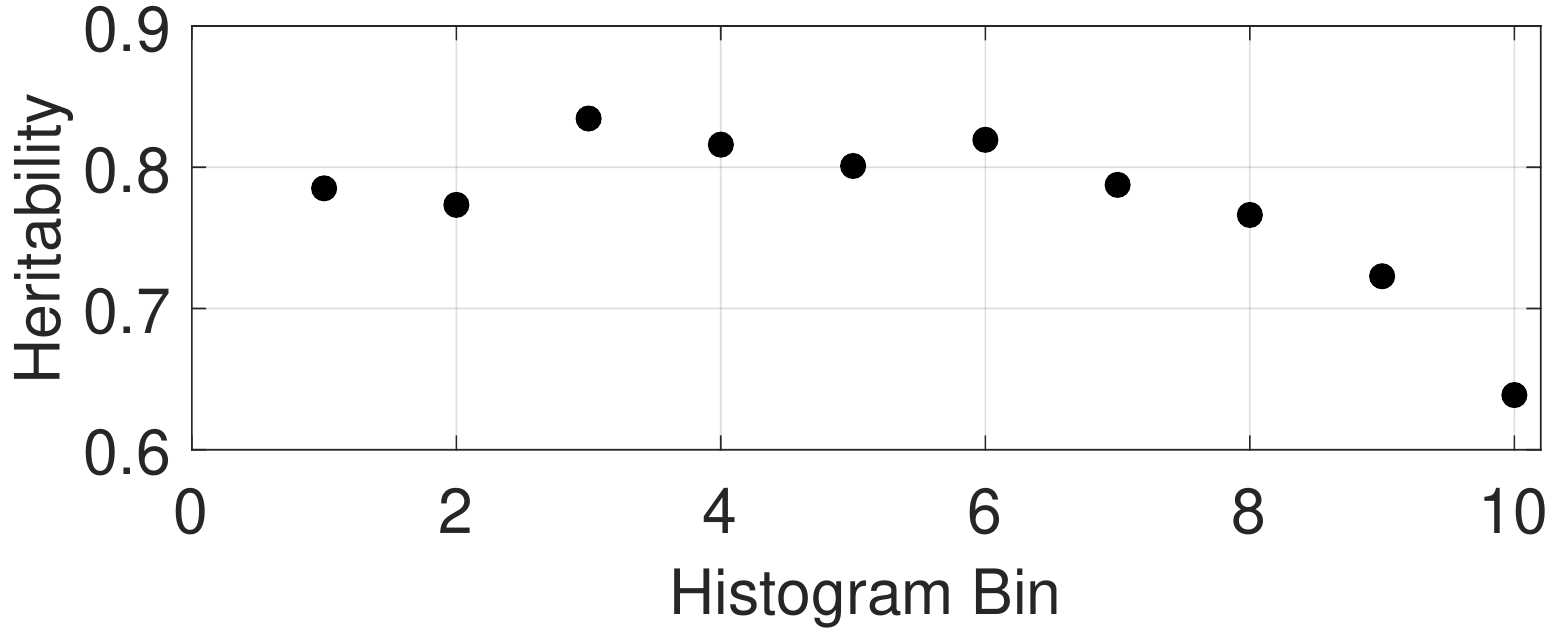}
    \end{center}
    \caption{Heritability estimates over our dataset for the first 10 bins of the BSH are quite high, ranging from 0.64 to 0.82.}
    \label{fig:heritability}
\end{figure}

\begin{figure*}[h]
	\centering
	\subfigure[]
	{\label{Figure:det1}
		\includegraphics[width=1.7in]{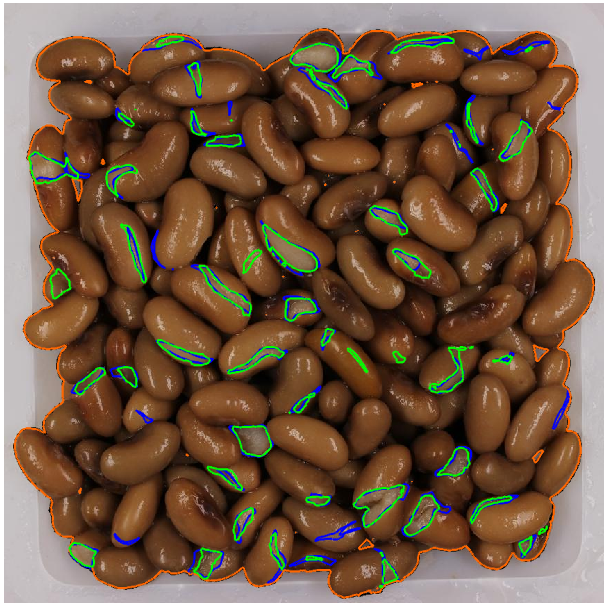}}
	\subfigure[]
	{\label{Figure:det2}
		\includegraphics[width=1.7in]{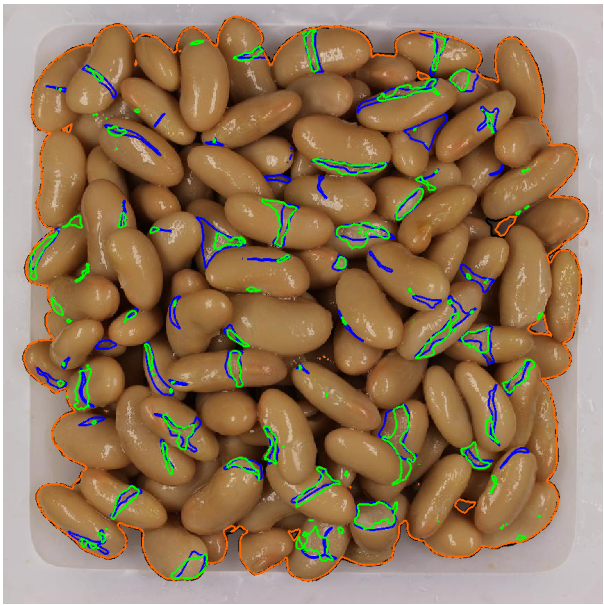}}
	\subfigure[]
	{\label{Figure:det3}
		\includegraphics[width=1.7in]{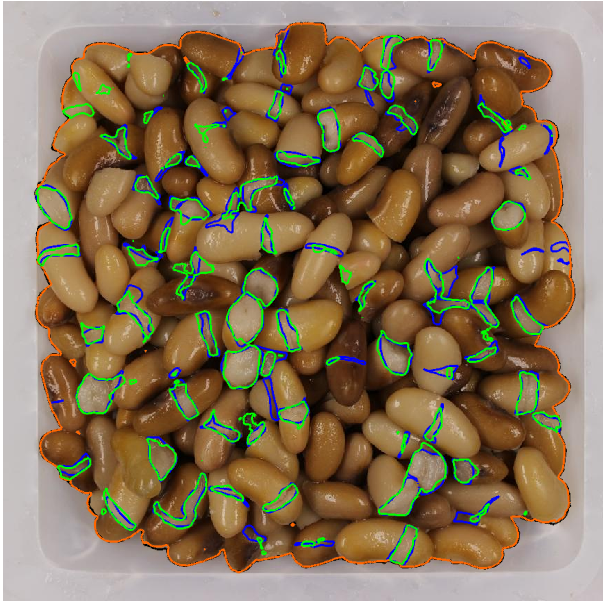}}
	\subfigure[]
	{\label{Figure:det4}
		\includegraphics[width=1.7in]{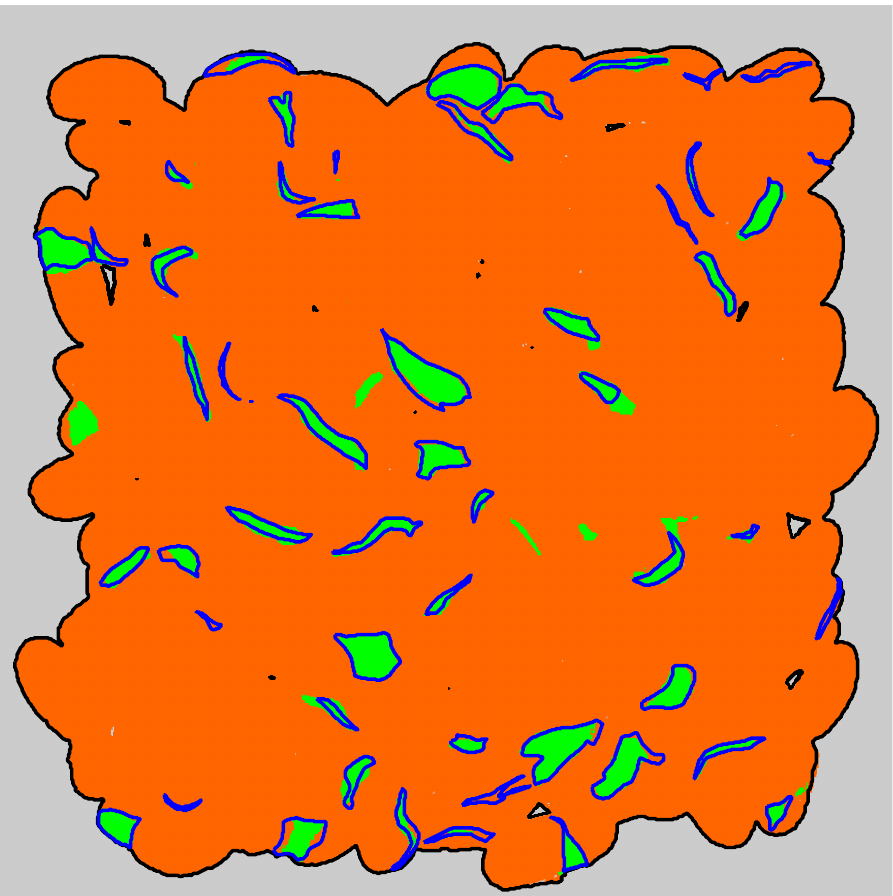}}
	\subfigure[]
	{\label{Figure:det5}
		\includegraphics[width=1.7in]{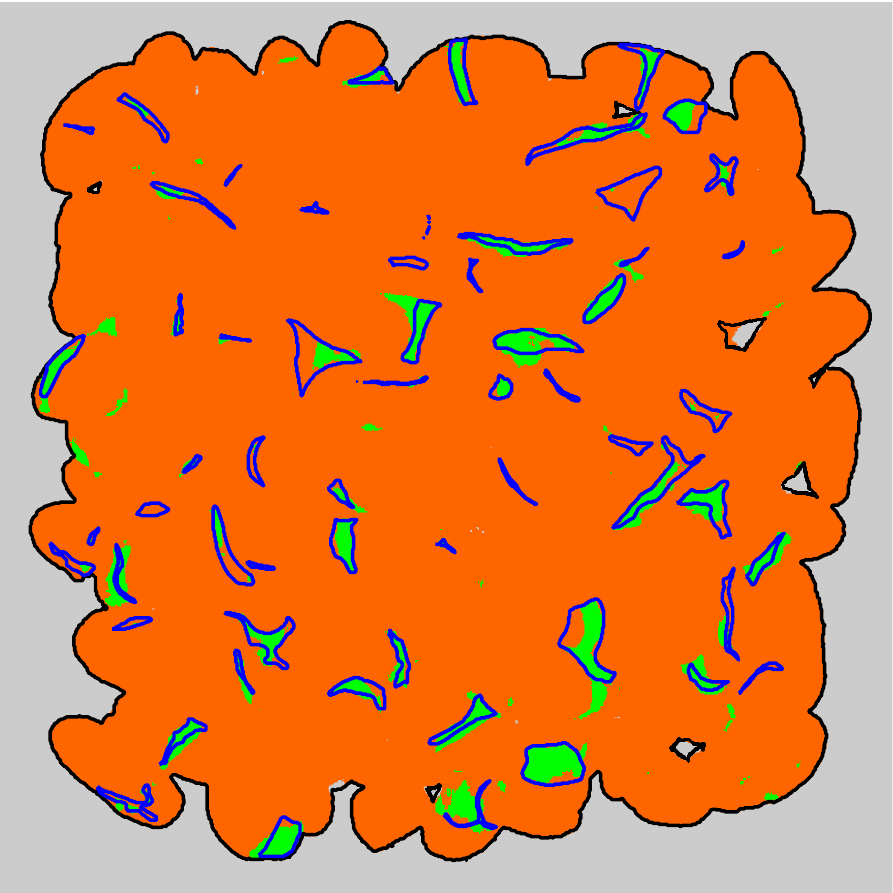}}
	\subfigure[]
	{\label{Figure:det6}
		\includegraphics[width=1.7in]{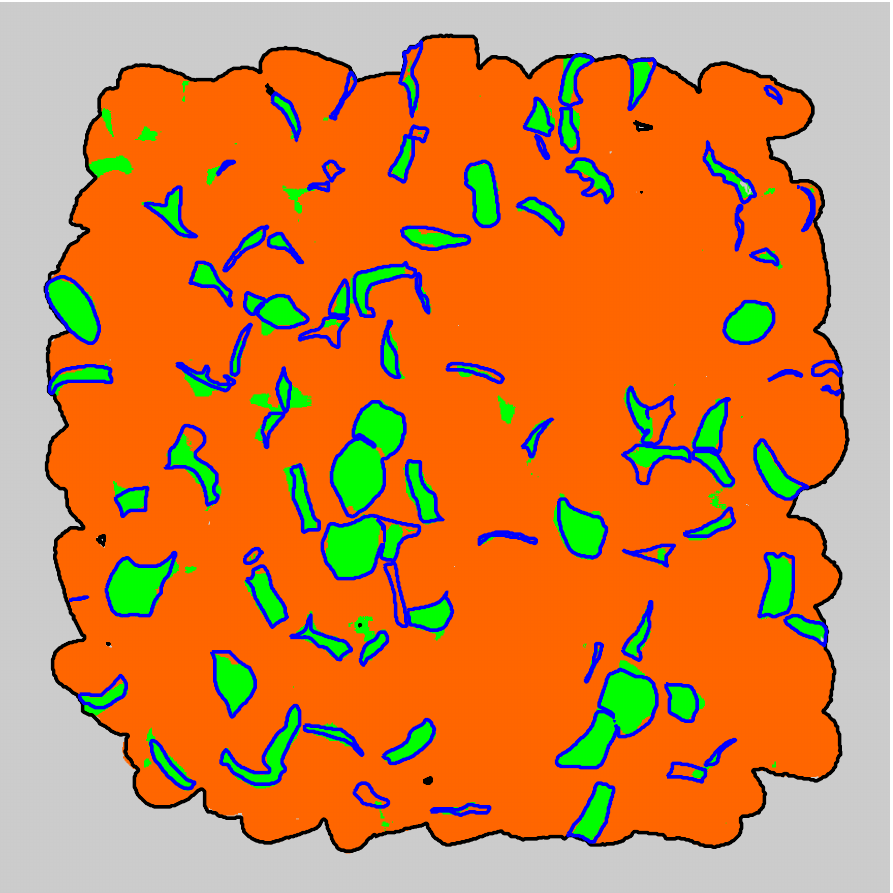}}
	\caption{Visualization of segmentation results: (a), (b) and (c) are images in the validation set; predicted (green) and true (blue) boundaries of splits are shown on top of the images. (d), (e) and (f) are visualization of segmentation of (a), (b) and (c), respectively, showing predicted tray (grey), predicted seed coat (orange) and predicted bean splits (green), along with true splits (blue) and bean regions (black) boundaries.}
	\label{Figure:detection}
\end{figure*}

Fig.~\ref{Figure:detection} shows segmentation results for 3 representative images in the validation set. Figs.~\ref{Figure:det1} to~\ref{Figure:det3} shows the boundaries of estimated and true splits plotted on the original images. Figs.~\ref{Figure:det4} to~\ref{Figure:det6} shows the predicted segmentation and ground truth. The model achieves higher accuracy on dark beans than light beans, as light bean seed coat color is very close to the split color. 

\subsection{Baseline performance comparison}
In Table~\ref{tab:error}, we use the bean pixels of the validation set to compare our method to a baseline Linear Discriminant Analysis (LDA)~\cite{martinez2001pca} on pixel-wise HSV color features~\cite{chen2008fast}.  The BSH error is quantified by Earth Mover's Distance~\cite{rubner2000earth}. The poor performance of LDA shows that pixel color alone is insufficient to detect splits and spatial cues learned by the  pyramid ConvNet greatly improve performance.

\begin{table}
	\begin{threeparttable}
		\caption{Baseline accuracy comparison for split detection}
        \label{tab:error}
        \begin{tabular}{lllll}
	    \toprule
     	\textbf{Methods} & \textbf{AP} & \textbf{IoU} & \textbf{BSR Error} & \textbf{BSH Error} \\
	    \midrule	
	    LDA~\cite{martinez2001pca}              & 0.30   & 0.26   & 33.2\%   & 3.75\\
	    Ours          & $\bm{0.82}$   & $\bm{0.58}$   & $\bm{9.6\%}$    & $\bm{0.81}$\\
	    \bottomrule
        \end{tabular}
	\end{threeparttable}
\end{table}

\subsection{Qualitative BSR and BSH results}

\begin{figure*}[h]
	\centering
	\subfigure[]
	{\label{Figure:dt1}
		\includegraphics[width=1.7in]{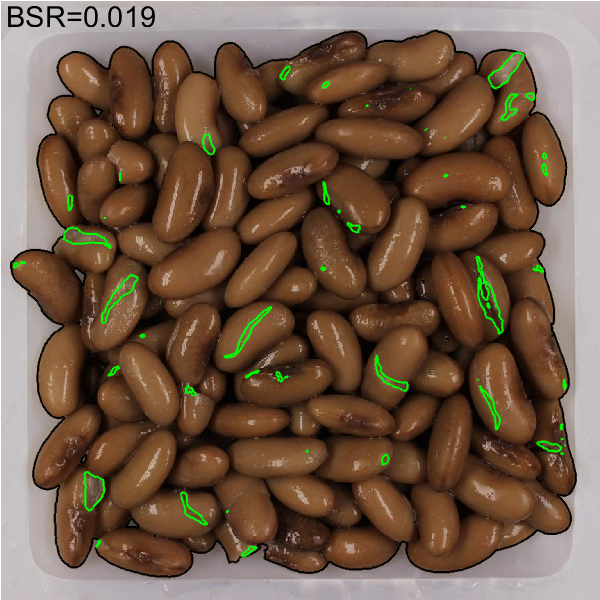}}
	\subfigure[]
	{\label{Figure:dt2}
		\includegraphics[width=1.7in]{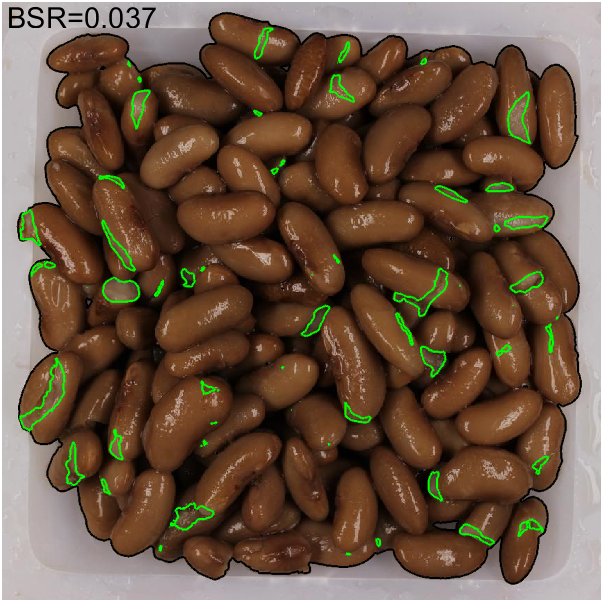}}
	\subfigure[]
	{\label{Figure:dt3}
		\includegraphics[width=1.7in]{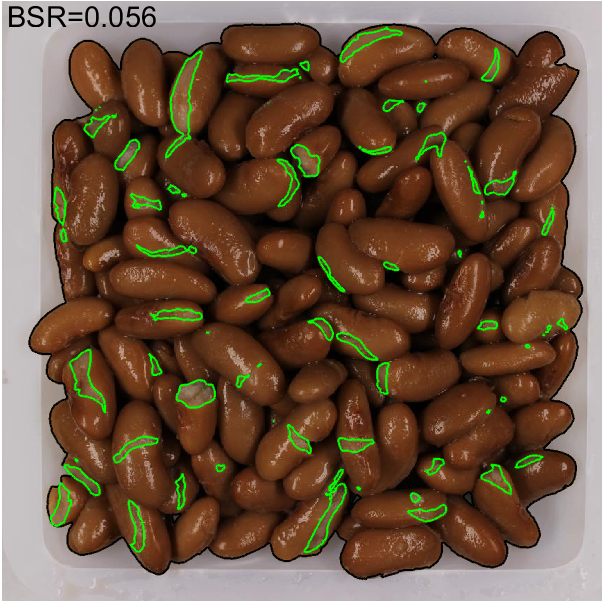}}
	\subfigure[]
	{\label{Figure:hist1}
		\includegraphics[width=1.7in]{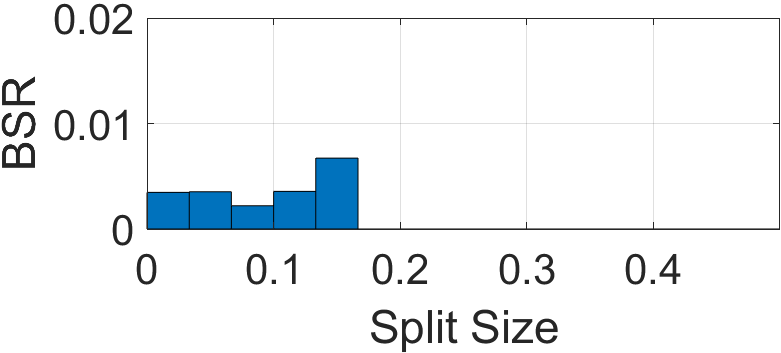}}
	\subfigure[]
	{\label{Figure:hist2}
		\includegraphics[width=1.7in]{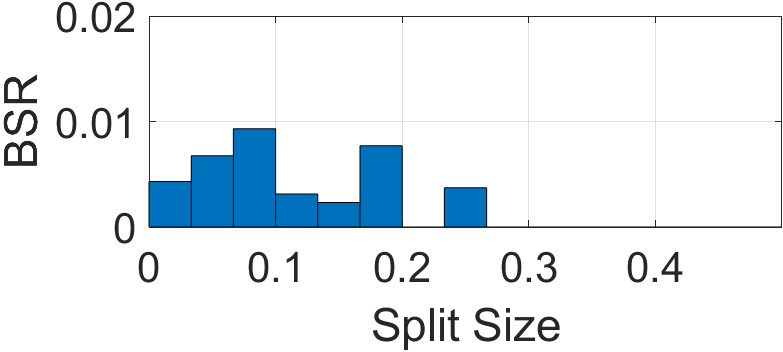}}
	\subfigure[]
	{\label{Figure:hist3}
		\includegraphics[width=1.7in]{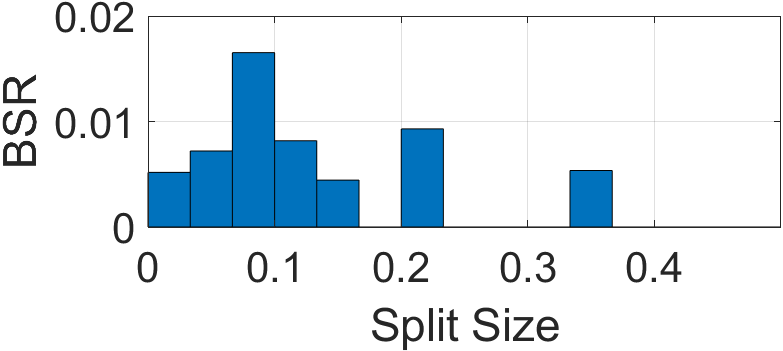}}
	\caption{(a) to (c) are 3 bean samples with increasing split severity due to increased retort time.  Our detected splits segments are shown by green boundaries and are quantified by BSR scores: (a) 0.019, (b) 0.037 and (c) 0.056.  The corresponding BSH measures are shown below in (d) to (f) with split size in units of maximum bean size.}
	\label{Figure:split_time}
\end{figure*}

Figs.~\ref{Figure:dt1} to~\ref{Figure:dt3} shows a single bean type under increasing retort times.  This results in more visible splits, and can be quantified by the increasing BSR measures. Figs.~\ref{Figure:hist1} to~\ref{Figure:hist3} show the BSH of Figs.~\ref{Figure:dt1} to~\ref{Figure:dt3}, respectively. To obtain the size of individual split, we cluster split pixels by using connected component labeling~\cite{rosenfeld1966sequential}. Note that different splits that are connected in images can be incorrectly treated as one split.

\subsection{Comparison of automated and human analysis}

Multiple human panelists rated each tray of beans with an integer intactness score ranging from 1 to 5. Their average score was used as reference intactness.  We calculated the Pearson correlation between the image-based BSR and the reference intactness, and compared this to the correlation of the individual raters with the reference intactness.\footnote{In correlating each  rater with the reference intactness, we excluded that rater's score from the intactness measure.}  BSR had a correlation coefficient of $r = -0.732$ compared to an average $r = 0.636$ over all raters. Thus the BSR tracks the reference intactness better and so has potential to achieve at least comparable performance to human panelists in intactness estimation.

\subsection{Applicability of methods to breeding}

The most important quality of detected features for bean breeders is a measure of the proportion of variance attributed to genotype rather than other factors. This estimate is called the repeatability or heritability, and ranges from 0 to 1, with higher values being more useful and effective to implement in a breeding program. The values indicate to what degree the phenotype is controlled by genotype, which determines how closely the progeny will resemble their parents, and therefore how effective selection based on the given trait (or features) will be from one generation to the next. Heritability estimates for the BSR was 0.957, which is extremely high. Heritability estimates for individual BSH bins representing different sizes of splits ranged from 0.64 to 0.82 (see Fig.~\ref{fig:heritability}), and some of these bin scores were impacted by retort time, rep, or whether the image was included in the training set. Though a significant proportion of the variation in human ratings was also attributable to genotype (Table~\ref{tab:variance}), these values were also significantly impacted by the Reviewer and Retort Time, which was not the case for the BSR trait. In addition, the need for multiple reviewers to reduce bias in human scores translates to a very time- and cost-intensive rating process compared to an image-based approach.

\section{Conclusion}
In this paper, we present two measures, BSR and BSH, that quantify and characterize the severity of bean splits.  We propose a method estimating BSR and BSH from bean images and demonstrate it on a new, annotated bean-image dataset.  We find strong correlation between BSR and human-estimated bean intactness.  Furthermore, from our bean dataset we find strong heritability of these measures, indicating that they are likely a useful tool for breeders interested in selecting for improved canning quality.  Since these measures can be estimated more quickly and reliably than human-based scores, they may improve breeders' ability to select genotypes for crossing and advancement in their breeding programs, as well as determine the genetic factors controlling propensity for seed coats to split during canning.

\section*{Acknowledgements}
This work was made possible in part by a predoctoral training award to Yunfei Long from grant number DGE-1828149 from the National Science Foundation. The authors would like to thank Kirk Dolan in the Food Science and Human Nutrition Department at  Michigan State University for providing equipment and technical support for the bean canning. 

\section*{Appendix}
\subsection*{A.1. Bean Genotypes}
\label{sec:beans}
Two yellow dry bean genotypes Ervilha (ADP0512) and PI527538 (ADP0468) were crossed, and the progeny were self-pollinated through single-seed descent for five generations to create recombinant inbred lines. Recombinant inbred lines are highly homozygous genotypes that incorporate various combinations of parental genetic material due to multiple recombination events. The parents Ervilha and PI527538 were selected from the Andean Diversity Panel \cite{Cichy2015} for their cooking and nutritional characteristics.  The genotypes included in this study were selected to evaluate whether their canning quality differs across different retort processing times.  The dataset contains images of both parents as well as 18 recombinant inbred lines canned in a retort across five different process times (10, 15, 20, 30 and 45 minutes).

\subsection*{A.2. Canning protocol and quality evaluation}
Each genotype was processed in duplicate across five retort times for a total of ten samples per genotype. For each sample, 90g dry weight (14-17\% moisture) of dry bean seeds were placed in mesh bags and soaked for 12 hours in 0.0028\% CaCl\textsubscript{2} solution prior to canning. The soaked samples were placed into aluminum cans, filled with brine (1.5\% sucrose, 1.25\% NaCl, 0.03\% CaCl\textsubscript{2}), and heated in an exhaust box to 165 \degree C. The cans were then sealed and processed at 250 \degree F in a stationary rotary retort for 10, 15, 20, 30, or 45 minutes. All process times exceeded minimum safety requirements (F\textsubscript{0}\textgreater 6 min) for production of canned food products.  Cans were left to equilibrate at room temperature for one week before opening, at which point visual inspections were performed. Each can was poured into a paper food tray and mixed slightly to distribute the samples evenly across each tray. The samples were evaluated by 14 trained reviewers (with at least 7 reviewers per can) using a 5-point scale for intactness ( 1: 0-20\% intact, 2: 21-40\% intact, 3: 41-60\% intact, 4: 61-80\% intact, 5: 81-100\% intact). Intactness is defined as an absence of splits. Following visual evaluations, each sample was rinsed and imaged.

Analysis of variance of intactness data was evaluated using the residual maximal likelihood method.  The model included genotype, retort time, genotype by retort time, replication, reviewer, and genotype by reviewer.  Genotype and retort time were fixed effects and replication and reviewer were random effects.

{\small
\bibliographystyle{ieee}
\bibliography{MyRef}
\small}

\end{document}